\documentclass[10pt,twocolumn,letterpaper]{article}

\usepackage{wacv}
\usepackage{times}
\usepackage{epsfig}
\usepackage{graphicx}
\usepackage{amsmath}
\usepackage{amssymb}
\usepackage{makecell}
\usepackage{pdfpages}
\usepackage{multirow}

\graphicspath{ {./images/} }

\usepackage[pagebackref=true,breaklinks=true,colorlinks,bookmarks=false]{hyperref}
\wacvfinalcopy 

\def\wacvPaperID{922} 

\ifwacvfinal\fi
\begin{document}

\title{FlowNet3D++: Geometric Losses For Deep Scene Flow Estimation}

\author
{
    Zirui Wang\\
    {\tt\small ryan@robots.ox.ac.uk}
    \and
    Shuda Li\\
    {\tt\small shuda@robots.ox.ac.uk}
    \and
    Henry Howard-Jenkins\\
    {\tt\small henryhj@robots.ox.ac.uk}
    \and
    Victor Adrian Prisacariu\\
    {\tt\small victor@robots.ox.ac.uk}
    \and
    Min Chen\\
    {\tt\small min.chen@oerc.ox.ac.uk}
}

\maketitle
\ifwacvfinal\thispagestyle{empty}\fi

\begin{abstract}
We present FlowNet3D++, a deep scene flow estimation network. Inspired by classical methods, FlowNet3D++ incorporates geometric constraints in the form of point-to-plane distance and angular alignment between individual vectors in the flow field, into FlowNet3D~\cite{liu2019flownet3d}. We demonstrate that the addition of these geometric loss terms improves the previous state-of-art FlowNet3D accuracy from \textbf{57.85}\% to \textbf{63.43}\%.
To further demonstrate the effectiveness of our geometric constraints, we propose a benchmark for flow estimation on the task of dynamic 3D reconstruction, thus providing a more holistic and practical measure of performance than the breakdown of individual metrics previously used to evaluate scene flow. This is made possible through the contribution of a novel pipeline to integrate point-based scene flow predictions into a global dense volume. FlowNet3D++ achieves up to a \textbf{15.0\%} reduction in reconstruction error over FlowNet3D, and up to a \textbf{35.2\%} improvement over KillingFusion \cite{slavcheva2017killingfusion} alone. We will release our scene flow estimation code later.


\end{abstract}

\section{Introduction}

Scene flow is defined as a 3D vector field that provides a low level representation of 3D motion. It is analogous to optical flow, which describes the pixel movements on a 2D image plane. Optical flow can be considered the projection of scene flow into 2D. Applications such as object detection, object tracking, point cloud registration, correspondence estimation and motion capture can benefit from this low-level information for better performance. 

Although the vector field representation is simple, scene flow estimation is far from an easy task. This is due to the requirement of accurate depth estimation and also the need to deal with occlusion. Traditionally, scene flow estimation is computed by optimising photometric error \cite{liu2019flownet3d}, or through matching hand-crafted features ~\cite{Brox_PAMI10_OpticalFlow}, each applied over multiple view geometry or RGB-D images. With the fast development of deep learning, some works bring CNN to scene flow estimation. This allows for the scene flow estimation to benefit from the semantic information and powerful deep feature extraction. Recently, PointNet~\cite{PointNet_Charles2017} and PointNet++~\cite{PointNet2_Qi2017} enabled the direct point cloud processing for deep learning. These works are particularly interesting since they are point cloud-based networks that can directly process the rich 3D geometric information, rather than implicitly learning 3D geometry from 2D images. Built on top of PointNet++, FlowNet3D \cite{liu2019flownet3d} tackles the scene flow estimation problem on point cloud directly, achieving the state-of-art scene flow estimation results. Despite the impressive results of FlowNet3D, point cloud based scene flow estimation is still at the very beginning of its development. FlowNet3D trains the network with a naive supervision signal, which is the $L_2$ loss between predicted flow and ground truth vectors.

In this work, we apply geometric principles from classical point cloud registration algorithms in order to mature deep scene flow estimation beyond the simple $L_2$ norm between prediction and ground truth.
In particular, we investigate two geometric constraints including: 1) point-to-plane distance and 2) the cosine distance between predicted flow vector and ground truth vector. The point-to-plane distance is a common loss term in Iterative Closest Point (ICP) \cite{besl1992method,newcombe2011kinectfusion}  algorithm, which is known for fast convergence. Cosine distance can penalise the angle between two vectors directly. As a result of this, cosine distance encourages correct alignment of our predicted scene flow vectors, not only that they lie on the $L_2$ norm surface.
The application of these geometrically principled constraints not only demonstrates improved accuracy over the state-of-art, but also improved convergence speed and stability of training.

Further, we introduce a novel benchmark for investigating the practical performance of scene flow estimators, through the proxy task of dynamic 3D reconstruction. Application of scene flow to dynamic reconstruction provides a holistic combination of the individual metrics previously used to evaluate 3D flow estimation. We contribute a novel pipeline for integrating point-based scene flow into a global dense volume.

To summarise, our main contributions are:
    \begin{enumerate}
        \item We improved the state-of-art point cloud-based scene flow estimation accuracy from 57.85\% to 63.43\% by combining point-to-plane loss, cosine distance loss with $L_2$ loss.
        \item We propose an average angle error metric to evaluate the flow direction deviation to supplement End Point Error (EPE), which is not sufficient alone to evaluate the angle difference between two vectors.
        \item We propose dynamic 3D reconstruction, alongside a novel dynamic integration pipeline, as a benchmarking task for scene flow estimation. 3D reconstruction provides a holistic and inherently geometric measure of flow estimation performance. Within our deformable scene flow benchmark task, our FlowNet3D++ achieves up to 15.0\% less reconstruction error than FlowNet3D, and up to a 35.2\% improvement over KillingFusion. 
    \end{enumerate}

The remainder of the paper is organised as follows: in Section~\ref{sec:related}, we briefly describe related work. Section~\ref{sec:method} describes our modifications to FlowNet3D, as well as our method to integrate the sparse scene flow vector field in dense dynamic reconstruction. In the experiment section, we evaluate the effect of adding different geometric constraints to scene flow estimation and our dynamic reconstruction results with several public datasets. Section~\ref{sec:conclusion} concludes our work and describes potential future work.

\section{Related Work}\label{sec:related}

\subsection{Tradition Scene Flow Estimation}
Scene flow is a low-level representation of 3D motion of points within a scene. It is a 3D extension from the 2D optical flow~\cite{Brox_PAMI10_OpticalFlow}, which itself describes the pixel movements on a 2D image plane. Many works have focused on estimating scene flow using multi-view geometry~\cite{vedula1999three} by associating salient image key points. Later works ~\cite{Pons2007,Huguet2007, Wedel2008,Vogel2011} tackle this problem with joint variational optimisation of image registration and motion estimation. \cite{Wedel2008} compute dense scene flow from stereo cameras and achieved 5fps on a CPU.  SphereFlow~\cite{SphereFlow_Hornacek2014,PD_FLOW_Jaimez2015} is the first real-time scene flow estimation system using RGB-D input. \cite{MC_FLOW_Jaimez2015} proposed to process rigid and non-rigid segments differently.

\subsection{Deep Flow Estimation}
The recent development in deep neural networks provides an alternative to address the problem of associating points over deformed depth maps. One group of deep methods can be viewed as the successors of the classic 2D optical flow methods. For example, FlowNet~\cite{Dosovitskiy_ICCV15_FlowNet} and its variants~\cite{ilg2017_flownet2}. Instead of using hand-crafted feature for tracking pixel locations, these methods rely on learned deep features for tracking and then back-project into depth maps to fetch the 3D scene flow. For better training and evaluation, Mayer et al \cite{mayer2016large} created three synthetic scene flow datasets. They also proposed a network for disparity and scene flow estimation. \cite{Cascaded_Ren2017} assume a dynamic scene contains foreground objects and background and apply instance segmentation masks over foreground to treat foreground and background differently. 

Similarly, \cite{Shao2018} developed a neural network that jointly estimates object segmentation, trajectories of objects, and the object scene flow from two consecutive RGB-D frames. Ilg et al. \cite{Occu_ILG} proposed a network based on FlowNet \cite{Dosovitskiy_ICCV15_FlowNet} to estimate occlusions and disparity together. \cite{Deep_Rigid_Instance_Ma2019} integrates three vision cues to estimate scene flow for rigid objects in self-driving tasks. The three vision cues are segmentation masks, disparity map, and optical flow and they are extracted by existing networks, i.e. Mask R-CNN \cite{MaskRCNN}, PSM-Net \cite{PSM_Net}, and PWC-Net \cite{PWC_Net}. 

All the above approaches are mostly image-based so that appearance features can be conveniently extracted using 2D convolution. However, some sensory data such as laser scanners is unstructured and therefore conventional convolution is not applicable. To address the problem, Behl et al \cite{PointFlowNet} evaluated the performance of scene flow estimation when integrating bounding box and integrating pixel-wise segmentation to scene flow estimation pipeline. \cite{PointFlowNet, Flownet3d, HPLFlowNet_Gu2019} are designed for scene flow estimation on the point cloud. PointFlowNet \cite{PointFlowNet} proposed to estimate scene flow, ego-motion and rigid object motion at the same time. In comparison to the PointFlowNet, FlowNet3D \cite{Flownet3d} and HPLFlowNet \cite{HPLFlowNet_Gu2019} are more general scene flow estimation frameworks that do not rely on rigid object assumption. More specifically, FlowNet3D extracts features with PointNet++ \cite{PointNet2_Qi2017}, mixes features and computes a coarse scene flow using a flow embedding layer, and propagates coarse scene flow to finer level using a set-upconv layer. HPLFlowNet, instead of using PointNet++, states that using permutohedral lattice\cite{Permutohedraw_lattice} and Bilateral Convolutional Layer (BCL) \cite{BCL} can improve global information extraction and faster performance. 

\section{Method} \label{sec:method}

\begin{figure*}[!t]
    \centering
    \includegraphics[width=0.9\textwidth]{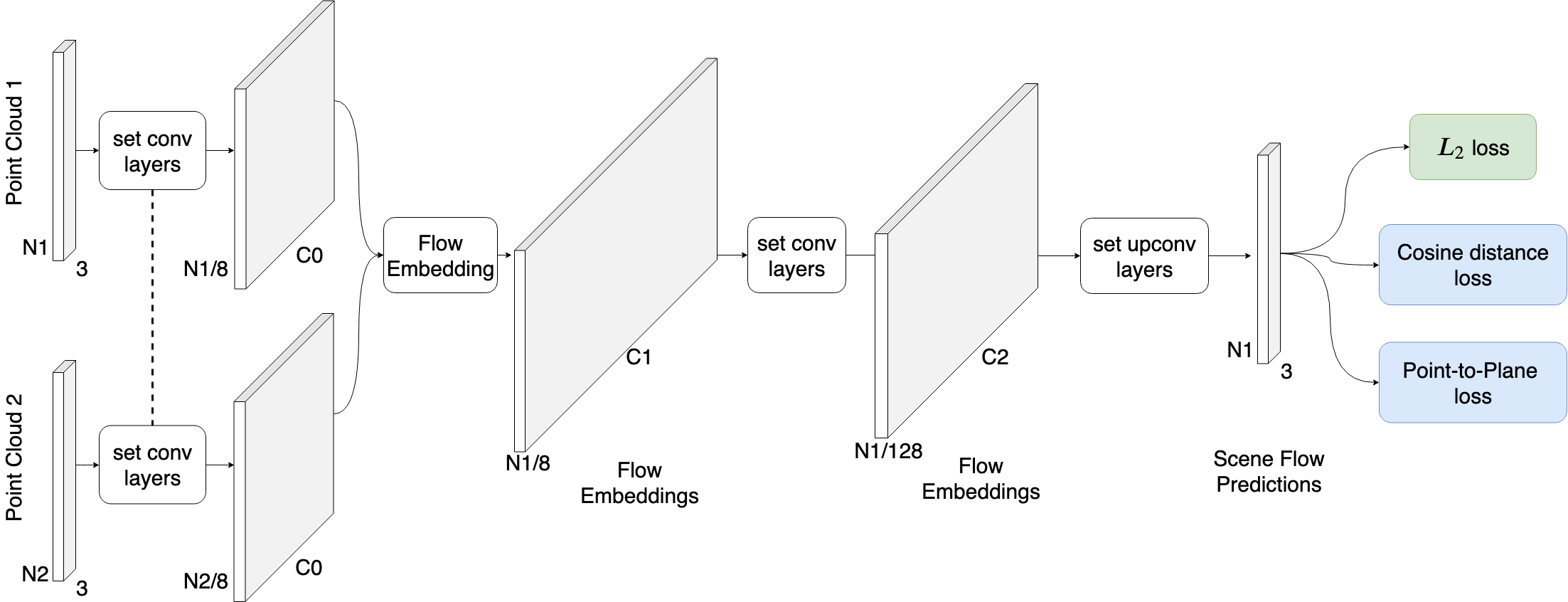}
    \caption{We adapt FlowNet3D structure and highlight the loss terms we added with blue boxes.} 
    \label{fig:flownet3d++}
\end{figure*}

FlowNet3D is a neural network for estimating 3D scene flow $\mathcal{V}_{ts}$ given two point clouds, namely the source point cloud  $\mathcal{X}_s$ and the target point cloud $\mathcal{X}_t$, where $\mathcal{X}_s$ and $\mathcal{X}_t$ are two sets of unordered 3D points. For generality, the numbers of both point clouds do not have to be identical, i.e. $|\mathcal{X}_s|\neq|\mathcal{X}_t|$, but the predicted vector field always has the same dimension as the source point cloud. FlowNet3D adopts the Siamese architecture that first extracts down-sampled point features for each point cloud using the PointNet++, and then mixes the features in the flow embedding layer. In the end, the output features of the flow embedding are imposed with the regularisation and up-sampled into the same dimensionality as the $\mathcal{X}_s$. The network is trained using the loss function of the $L_2$ norm between predicted $\|\mathcal{V}_{ts}-\mathcal V_{gt}\|_2$ where $\mathcal V_{gt}$ is the groundtruth scene flow field.

FlowNet3D has been successfully applied in rigid scenes. In this paper, we further explore the potential of it when applied to non-static scenes, even the scenes dominated by deformable objects. More importantly, we introduce two loss terms that improves the accuracy of the prediction in both dynamic scenes while maintains the performance in rigid scene (measured using KITTI dataset).  The new loss term also speeds up and stabilises the training procedure. Fig \ref{fig:flownet3d++} illustrates the general idea of FlowNet3D and loss terms we applied. More details on the original FlowNet3D structure can be found in its paper.

\subsection{Geometric Constraints}
\textbf{Point-to-Plane Loss} is inspired by the popular point-to-plane distance metric for point cloud registration, such as the Iterative Closest Point (ICP) algorithm. Specifically, we can use the set $\mathcal{X}_l^n$ and $\mathcal{X}_w^n$ to represent two 3D point clouds at the $n^{th}$ frame, where the labels $l$ and $w$ represent the live camera and the world coordinate system, respectively.  Each point $\mathbf{x}_\in\mathcal{X}_w^n$ is a 3D homogeneous coordinate. The point $\mathbf{x}_w^n$ can be transformed from the world coordinate into the camera coordinate using $\mathbf x_l^n =\mathbf T_{cl}^n \mathbf x_w^n$, where $\mathbf T^n_{lw} \in \mathbb{SE}(3)$ is a 3D rigid transformation.

Given $\mathcal{X}_l^n$ and $\mathcal{X}_w^n$, $\mathbf T^n_{cw}$ can be estimated by minimising the following error function from a typical ICP algorithm~\cite{newcombe2011kinectfusion}: 
    \begin{equation}
        \mathbf T^n_{cw} = \arg\min_{\hat{\mathbf T}^n_{cw}} \sum_{\mathbf{x}_w^n \in \mathcal{X}_w^n}\|\mathbf n(\mathbf x^n_c)^\top (\mathbf T^n_{cw} \mathbf x^n_w - \mathbf x^n_c ) \|^2, \label{eq:icp}
    \end{equation}
where $\mathbf{n}(\mathbf x^n_c)$ is the function to calculate the surface normal at $\mathbf x^n_c$. $\mathbf x^n_c$ is the closet point to  $\mathbf T^n_{cw} \mathbf x^n_w $. The dot product between the surface normal and the closet distance measures the distance from $\mathbf T^n_{cw} \mathbf x^n_w$ to the plane defined by $\mathbf x^n_c$ and its normal, hence it is known as the point-to-plane metric. 

Inspired by the point-to-plane metric, we introduce a new loss for training the FlowNet3D, which is defined as follows:
    \begin{equation}
        \mathcal{L}_{pp} = \sum_{\mathbf{x}_s \in \mathcal{X}_s}\|\mathbf n(\mathbf x_t)^\top (\mathbf x_s - \mathbf x_t ) \|^2, \label{eq:pt_plane_loss}
    \end{equation}
where $\mathbf{x}_t$ is the closest point in the target set $\mathcal X_t$ to the source point $\mathbf x_s\in\mathcal{X}_s$. The scene flow may encode any rigid transformation or simply the non-static motion field, which is ultimately determined by the the samples provided during training. During training on FlyingThings, both $\mathcal X_s$ and $\mathcal{X}_t$ are in the same coordinate system and therefore, the trained model naturally learns to represent segments of rigid motion fields. Interestingly, we found that the same model can generalise to the point clouds extracted from consecutive frames of a deforming object so well that it outperforms the state-of-the-art dynamic fusion algorithm. 

\begin{figure}[ht]
    \centering
    \includegraphics[width=0.5\linewidth]{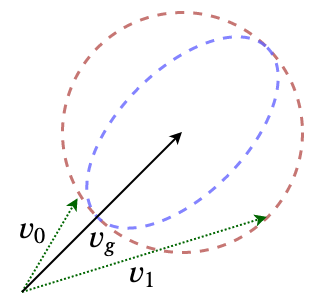}
    \caption{The red circle denotes the energy contour that $\mathcal{L}_2$ loss penalises equally. The blue circle denotes the energy contour after combining $\mathcal{L}_2$ and $\mathcal{L}_{\cos}$. $\mathcal{L}_{\cos}$ explicitly punishes vectors with large angle deviations. In this figure, $v_1$ is penalised more than $v_0$ after adding $\mathcal{L}_{\cos}$.
    }
    \label{fig:cos_plus_L2}
\end{figure}

\textbf{Cosine Distance Loss} aims at constraining the angle between predicted flow field and the ground truth. From the scene flow predictions of FlowNet3D, we noticed that some of the predicted motion vectors differ greatly in direction from the groundtruth. As a result, we introduce the cosine distance loss which aims to minimise the angle between prediction and ground truth. We compute the cosine distance directly between a predicted vector and its groundtruth. This provides extra penalisation to vectors with directions with deviate from the groundtruth, even if they have the same $\mathcal L_2$ loss. Fig \ref{fig:cos_plus_L2} illustrates the effect of applying $\mathcal L_2$ loss and cosine distance together. 

\textbf{Combined Loss} includes all three different loss terms using a weighted summation:
\begin{equation}
    \mathcal{L} = \mathcal{L}_2 + \lambda_{p}\mathcal{L}_{pp} + \lambda_{\cos}\mathcal{L}_{\cos}
\end{equation}
where $\lambda_p$ and $\lambda_{cos}$ are the weight to balance among the loss terms, the $\mathcal{L}_2 =\frac{1}{|\mathcal V'_{ts}|}\sum_{\mathbf v\in\mathcal V'_{ts}}\|\mathbf{v}-\mathbf v_{gt}\|_2$ and $\mathcal L_{cos}=\frac{1}{|\mathcal V'_{ts}|}\sum_{\mathbf v\in\mathcal V'_{ts}} \frac{\mathbf v\cdot \mathbf v_{gt}}{\|\mathbf v\|\|\mathbf v_{gt}\|}$. The $\mathcal V'_{ts}$ is the predicted vector field and $\mathbf v_{gt}$ is an individual ground truth vector corresponding to $\mathbf v$. It is worth noting that the cosine loss and $\mathcal L_2$ loss weigh over the angles and lengths of the predicted vector field, respectively.

\subsection{Scene Flow for Dynamic 3D Reconstruction}
The performance of flow field estimation on a dynamic scene is normally evaluated through the counting of inliers, which are determined through a set threshold. However, this evaluation scheme depends heavily on the threshold, which must be set heuristically. We propose to benchmark scene flow framework based on a state-of-the-art dynamic 3D reconstruction system, so that the scene flow can be evaluated by viewing a 3D model. This provides a more holistic performance measure, as well as a practical application for 3D flow estimation. 

Dynamic 3D reconstruction is recently introduced for recovering non-static objects, including deformable objects such as moving animals or human beings~\cite{newcombe2015dynamicfusion,innmann2016volumedeform,slavcheva2017killingfusion,slavcheva2018sobolevfusion,joo2015panoptic,joo2018total}.  KillingFusion~\cite{slavcheva2017killingfusion} and its variant SobolevFusion~\cite{slavcheva2018sobolevfusion} represent the state-of-the-art dynamic fusion method directly estimating a dense vector field between two TSDF volumes. However, this variational optimisation process is easily trapped in local minimum when the search space is large. Our benchmark framework, which can also be considered as a dynamic reconstruction system, significantly outperforms KillingFusion in terms of quality by a 35.2\% reduction in mean error.

Particularly, our benchmark framework takes in a sequence of point clouds with corresponding scene flow predictions to recover a 3D model. The reconstruction error can be visualised comparing with the ground truth model.
In experiments, the FlowNet3D++ reduces up to 15.0\% error in the dynamic reconstruction task compare to FlowNet3D.

\textbf{Overview} of the whole pipeline is illustrated in Fig.\ref{fig:flownet3d++} and we show the essential steps below:
    \begin{enumerate}
        \item Compute the rigid $\mathbb {SE}(3)$ transformation between the live point cloud and the canonical model using point-to-plane ICP algorithm~\cite{besl1992method} as shown in Eq.~(\ref{eq:icp}). It compensates the overall movement of the target object by absorbing it into the current camera pose. 
        \item\label{s0}Predict the scene flow between the live and the canonical point cloud. In this paper we experimented the FlowNet3D\cite{liu2019flownet3d} and our FlowNet3D++.
        \item\label{s1} Warp the live point cloud using the scene flow computed in the last step and create the synthetic depth map by projecting the warped live point cloud into the compensated camera pose. 
        \item\label{s2} Construct a live TSDF volume $\phi_n$ from the synthetic depth map using the widely used depth to volume integration method, which is first introduced in KinectFusion \cite{newcombe2011kinectfusion}.
        \item\label{s3} Refine the vector field $\Psi_V$ between the live volume and canonical volume using a simple variational voxel based scene flow refinement.
        \item Update the $\phi_{global}$ by taking the voxel-wise weighted average between $\phi_{n}$ and $\phi_{global}$ for the TSDF values live and accumulate the weight~\cite{curless1996volumetric}.
    \end{enumerate}
Step \ref{s0} introduces the deep scene flow to warp the live point cloud so that a virtual TSDF volume that is much easier for the KillingFusion to optimise and therefore reduces the computation complexity and quality of the recovered model. Step \ref{s1}, \ref{s2} and \ref{s3} formulate our novel scene flow integrator that integrates a scene flow in point cloud resolution to the full TSDF volume resolution with very little artefacts. 

\textbf{Scene Flow Integrator} merges multiple point cloud into a single 3D volumetric representation from which the 3D model can be extracted. Specifically, assuming $\mathcal X_l^n$ represents point cloud of live frame in camera coordinate and $\mathcal X_g^n$ represents the ray-casted point cloud from the canonical model, the scene flow predictor computes a scene flow field $\Psi^n_S$ that associate $\mathcal X_l^n$ with $\mathcal X_g^n$. The warping from $\mathcal X_l^n$ to $\mathcal X_g^n$ can be formulated as follows:
\begin{equation}
    \mathcal X_g^{n\prime} =\{\mathbf x| \mathbf x := v(\mathbf x^n_l) + \mathbf x^n_l\},
\end{equation}
where $\mathbf x^n_l\in\mathcal X_l^n$.
Note that  $\Psi^n_S$  and $\mathcal X_l^n$ share the same resolution and $\mathcal X_g^{n\prime}$ and $\mathcal X_g^n$ are different. Therefore, our target is to integrate $\mathcal X_g^{n\prime}$ into the canonical volume $\phi_{global}$ smoothly.

Naively integrating $\mathcal X_g^{n\prime}$ into the global TSDF volume seems a reasonable solution, however, in our experiments we discover this causes significant artefacts. This is because scene flow  computed on point clouds is only capable of inferring the motion on the object surface or the zero level set. To deform a volumetric TSDF, the vector field has to cover the entire 3D region within the truncated area while maintaining the property of TSDF to be a precise level set function so that the artefacts are minimised.

Therefore, we tackle the problem by:
1) creating a synthetic depth map $\tilde{D}^n=\{\tilde{d}(u,v)|(u,v)\in\Omega\}$, where $\Omega$ is a set of pixel locations on the depth map, by projecting $\mathcal X_g^{n\prime}$ onto the depth map $\tilde{D}^n$. 
2) creating a synthetic live volume $\phi_n$ from $\tilde{D}^n$. This is equivalent to integrating a depth map to an empty TSDF volume.

By converting the deformed point cloud $\mathcal X_g^{n\prime}$ into a TSDF volume $\phi_n$, we have acquired a coarse alignment between the $\phi_n$ and $\phi_{global}$. The next step is to refine this coarse alignment with a simple variational vector field refinement.

\textbf{Voxel Based Vector Field Refinement}: The concept of running variational optimisation directly on TSDF volume was first introduced in the KillingFusion~\cite{slavcheva2017killingfusion} and simplified in SobolevFusion~\cite{slavcheva2018sobolevfusion}. It solves the vector field by evolving the source TSDF into target TSDF iteratively. This approach enjoys the advantage of being capable of dealing with topological changes but a drawback of this variational SDF evolution lies in that it can easily get trapped in some local minima. This is because it lacks explicit correspondences associating level set functions. By providing a good initial solution from our deep scene flow estimator, only a few iterations of voxel based vector field refinement is needed. Specifically, for a voxel at position $\mathbf{x}\in\mathcal V$ and a 3D vector $\mathbf{v}= v(\mathbf{x})$ associates with this voxel, our energy is simply defined as:
    \begin{equation}
    E(\Psi_V) = \frac{1}{2} \sum_{\mathbf{x}\in\mathcal V}\left(\phi_n(\mathbf{x}+\mathbf{v}) - \phi_{global}(\mathbf{x})\right)^2
    \end{equation}
where $\phi(\mathbf{x})$ represent TSDF value at voxel centre $\mathbf{x}$ and the energy can be optimised using gradient descent easily:
    \begin{equation}
    \mathbf v^{(k+1)} = \mathbf v^{(k)}- \alpha E'(\mathbf v^{(k)})
    \end{equation}
where $\mathbf v^{(k)}\in\Psi_V^{(k)}$ represents the vector field $\Psi^n_V$ at its $k^{th}$ iteration and $E'(\mathbf v)$ is the gradient with respect to the $\mathbf v$ and can be computed efficiently using the following calculus of variations:
    \begin{equation}
    E'(\mathbf v) = (\phi_n(\mathbf x+\mathbf v) - \phi_{global}(\mathbf x)) \nabla\phi_n(\mathbf x+\mathbf v),
    \end{equation}
where the $\nabla\phi_n(\mathbf x + \mathbf v)$ is the spatial gradient at the voxel position $(\mathbf{x} + \mathbf{v})$ in the live volume $\phi_n$.

It is worth noting that the vector field computed from above optimisation is only meaningful in local regions and the purpose is two-fold: (i) to register a roughly aligned live volume to canonical model; (ii) to remove artefacts introduced in the coarse non-rigid point cloud registration. Thanks to the quality of the deep scene flow estimator, we no longer require a regularisation term, such as those in KillingFusion and SobolevFusion.

The above energy will produce a scene flow vector for each SDF voxel. In general, the magnitude of this vector field should be small because the main evolution has already been compensated when warping the live point cloud to the canonical point cloud. As a result, with a small number of iterations, typically ranging from 3 to 70, we can mediate the artefacts and noise introduced in from scene flow predictor.

We are aware that having variational refinement may affect the deep scene flow benchmarking result. However, this variational refinement is necessary for complex tasks like dynamic reconstruction. If not present, the tracking can fail after a few frames due to the large accumulated error. 
To eliminate the effect of this variational refinement in benchmarking, we explicitly set a fixed iteration number for all experiments. For the Snoopy and Duck dataset, we use 30 iterations for all deep scene flow benchmarking. 

\section{Experiments}\label{sec:experiments}
In this section, we evaluate our modifications to FlowNet3D and validate their effectiveness quantitatively in two subsections. In the first subsection, we benchmark our FlowNet3D++ result using the existing scene flow datasets FlyingThings and KITTI, which are pre-processed and provided by FlowNet3D. For pre-processing details, we refer reader to the FlowNet3D supplementary material. We also provide a graph to analyse the time taken for our training to converge. In the second subsection, we quantitatively evaluate the performance of FlowNet3D++ in our novel dynamic reconstruction benchmark. This is performed on two reconstruction datasets (Snoopy and Duck), both of which are provided by KillingFusion\cite{slavcheva2017killingfusion}. Further qualitative results can be found in our supplementary material. 

To enable the point-to-plane loss term in Eq. \ref{eq:pt_plane_loss}, we also pre-compute per-point surface normal for the FlyingThings dataset but we do \textbf{not} use surface normals as input features.

Our model is trained from scratch using the training split of FlyingThings dataset and testing is performed on the test split. We directly transfer our model that was trained on FlyingThings to KITTI without any fine-tuning. For the dynamic reconstruction benchmark, we also directly deploy the model that was trained on FlyingThings dataset to the pipeline, again without fine-tuning. For hyper-parameters, in most experiments we use exactly same hyper-parameters the FlowNet3D used to show the effectiveness of our loss terms. For the best result we show in Table \ref{tab:flyingthings}, we trained 200 epochs.

\subsection{Metrics}
We report our results using 3D End-Point-Error (EPE) and an accuracy metric (ACC) with two thresholds. These three metrics are also used in FlowNet3D to provide fair comparison. We also propose the average angle deviation error (ADE) for this task for the evaluation of the predicted scene flow vectors' direction.
\textbf{EPE}: the EPE is the $L_2$ norm between an estimated flow vector and its ground truth vector.
\textbf{ADE}: we define the ADE as $\arccos(\frac{1}{N}\sum\cos(v_p, v_{gt}))$, where $v_p$ and $v_{gt}$ are predicted vector and its ground truth vector.

\subsection{FlyingThings Dataset}
\begin{table}[t]
    \begin{tabular}{c|c|c|c|c|c}
    \hline
    \thead{Input\\channels} & Model & \thead{ACC\\(0.05)} & \thead{ACC\\(0.10)} & EPE & \thead{ADE\\(degree)} \\ \hline
    \multirow{2}{*}{xyz}    & F3D    & 23.71\%    & 56.05\%   & 0.1705 & 22.83 \\ 
                            & F3D++    & \textbf{28.50}\%    & \textbf{60.39}\%   & \textbf{0.1553} & \textbf{20.78} \\ \hline
    \multirow{2}{*}{rgb}    & F3D    & 25.37\%    & 57.85\%   & 0.1694 & 22.58 \\
                            & F3D++    & \textbf{30.33}\%    & \textbf{63.43}\%   & \textbf{0.1369} & \textbf{21.14} \\ \hline
    \end{tabular}
\label{tab:flyingthings}
\caption{F3D is a shorthand for FlowNet3D. We evaluate our FlowNet3D++ in two input settings. Input setting \textit{xyz} means we only use point position as input features. Input setting \textit{rgb} means both point position and the colour features are fed into the network. For EPE and ADE, the lower the better.}
\end{table}

In FlowNet3D++, we apply both the cosine distance loss and point-to-plane loss alongside the original $L_2$ loss. The results listed in Table~\ref{tab:flyingthings} show that our modifications improve all metrics that we test. In fact, the geometric-only XYZ-FlowNet3D++ even outperforms RGB-FlowNet3D, which is allowed to incorporate colour information. We use $\lambda_{pp}=1.3$ and $\lambda_{cos}=0.9$ for this test, but we found $\lambda_{pp, cos} \in [0.5, 1.5]$ generates good results in the general case. As the FlowNet3D did not evaluate ADE, we compute FlowNet3D's ADE with the pre-trained model provided by~\cite{liu2019flownet3d}.

\subsection{KITTI Dataset}
As KITTI scene flow dataset only provides a colourless, LiDAR-scanned point cloud, we only show the results for geometry-only models.

\begin{table}[h]
    \begin{center}
        \begin{tabular}{c|c|c|c}
        \hline
        Model                      & Outlier & EPE    & ADE   \\ \hline
        F3D (with our eval script) & 7.53\%  & 0.3259 & 42.60 \\ \hline
        F3D++                       & \textbf{4.81}\%  & \textbf{0.2530} & \textbf{36.86} \\ \hline
        \end{tabular}
    \end{center}
\caption{KITTI scene flow benchmark. We report that our model has significantly lower error in all tested error metrics.}
\label{tab:kitti}
\end{table}

We propose a more simple evaluation procedure on the KITTI dataset than was used in~\cite{liu2019flownet3d}. 
Instead of cutting the KITTI point cloud into numerous chunks and having to deal with overlapping regions, we resize the KITTI dataset to the size of FlyingThings scenes, which is $x \in [-15m, +15m], y \in [-8m, +8m], z \in [0m, 35m]$, before feeding it to networks. Although this produces differing results than in~\cite{liu2019flownet3d}, we ensure a fair comparison by training both FlowNet3D and FlowNet3D++ on FlyingThings and transferring to our resized KITTI without fine-tuning. We report our results in table \ref{tab:kitti}.

\subsection{Dynamic Dense Reconstruction}
In this section, we demonstrate the effectiveness of FlowNet3D++ within our proposed dynamic dense reconstruction benchmark.

\subsubsection{Configuration}\label{sec:config}
Our depth-only dynamic reconstruction system is implemented on top of InfiniTAM~ ~\cite{prisacariu2017infinitam}, an open sourced RGBD dense SLAM system with modern CUDA support. The volume resolution is set as $256^3$ and voxel size 3 mm or 5 mm for all of our experiments. Specifically, we use 3mm for small scenes like \textit{Snoopy} and \textit{Duck} dataset~\cite{slavcheva2017killingfusion} and 5 mm for the VolumeDeform datasets~\cite{innmann2016volumedeform}. The truncated distance $\delta$ is set to $\pm4$ times of the voxel size. The step size $\alpha$ for optimiser is set to $0.1$. We also implement a SobolevFusion system for comparison (comparison images can be found in Appendix). Similar to the KITTI scene, applying FlowNet3D++ to videos that captured with different cameras requires the scene to be resized to the range of FlyingThings dataset, i.e. $x \in [-15m, +15m], y \in [-8m, +8m], z \in [0m, 35m]$. The choice of scaling factors depends on different voxel size, SDF volume size and camera intrinsics. However, in practice, we found a rough estimation of the scaling factors works well for all the experiments. In particular, for the \textit{Snoopy} sequence, the scaling factors are set as $s_x = 25, s_y = 25, s_z = 30$. The good results achieved through this resizing method in dynamic reconstruction provide evidence that the resizing in KITTI evaluation is also valid.

\subsubsection{Results}

\begin{figure}[t] 
    \centering
    \includegraphics[width=\linewidth]{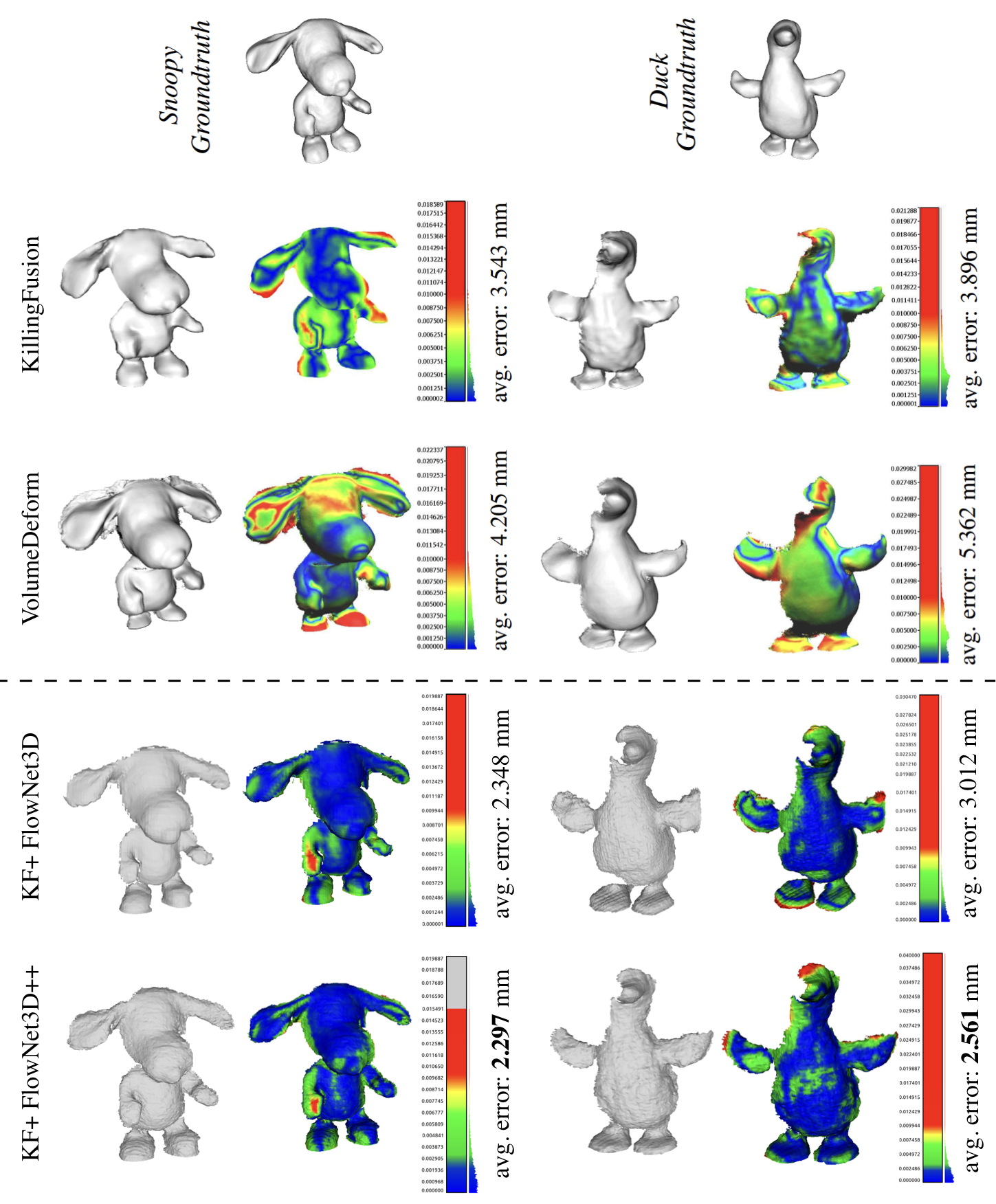}
    \caption{Mean mesh-to-mesh error. Blue regions represent low error. We use red to represent an error that is larger than 1cm. Top: ground truth mesh. $2^{nd}$ row and $3^{rd}$: results of KillingFusion and VolumeDeform (image copied from ~\cite{slavcheva2017killingfusion}). $4^{th}$ row: Our reconstruction results with unmodified \textit{FlowNet3D}. Bottom row: our reconstruction result after applying our \textit{FlowNet3D++}. It can be seen that most of our model appears as blue indicating low overall error. The bottom two rows demonstrate that FlowNet3D++ offers an improvement over FlowNet3D. Quantitatively, our average error on \textit{Snoopy} and \textit{Duck} scene is \textbf{35.2}\% and \textbf{34.3}\% lower than KillingFusion reconstructions, respectively.}
    \label{fig:duck-snoopy}
\end{figure}

The KillingFusion dataset (Snoopy and Duck) provides a ground truth mesh. Thus, we can quantitatively analyse the benefit of adding deep scene flow estimation to the dynamic reconstruction. We also present more images of running our systems on VolumeDeform dataset and a video sequence we record by ourselves in Appendix, to illustrate the benefit of our pipeline qualitatively. Our Snoopy and Duck evaluation result is reported in Table \ref{tab:err-comp} and Fig \ref{fig:duck-snoopy}.

\begin{table}[h]
\begin{center}
    \begin{tabular}{  c | c | c | c | c } 
    \hline
     & \multicolumn{4}{c}{Mean Error To Ground Truth (mm)} \\
    \hline
    Scenes & \thead{Volume\\Deform} & \thead{Killing\\Fusion} & \thead{KF +\\FlowNet3D} & \thead{KF +\\FlowNet3D++} \\ 
    \hline
    Snoopy & 4.205 & 3.543 & 2.348 & \textbf{2.297}\\ 
    \hline
    Duck   & 5.362 & 3.896 & 3.012 & \textbf{2.561}\\ 
    \hline
    \end{tabular}
\end{center}
\caption{Evaluation of FlowNet3D and FlowNet3D++ in our dynamic reconstruction benchmark}
\label{tab:err-comp} 
\end{table}

\section{Ablation study}
To validate the individual benefit derived from each of our geometric constraints, as well as their combination, we perform ablation tests for both the geometric-only models and colour models. Unless otherwise stated, we use exactly same training procedure as described in~\cite{Flownet3d}.

Results are shown in Table \ref{tab:ablation_xyz} and Table \ref{tab:ablation_rgb}. The results in the bottom rows of Table \ref{tab:ablation_xyz} and Table \ref{tab:ablation_rgb}, we trained for 200 epochs, instead of 150 epochs.

In addition to the overall performance of the geometric loss terms, it is also worth noting that
in the RGB setting, simply combining $\mathcal{L}_{pp}$ + $\mathcal{L}_{\cos}$ does not yield the best ACC and EPE after 150 epochs of training. Instead, the best result acquired after this schedule is the model trained with $\mathcal{L}_{pp}$.
However, we found that the accuracy of FlowNet3D + $\mathcal{L}_{pp}$ plateaus after 150 epochs, whereas the accuracy of model with both $\mathcal{L}_{pp}$ and $\mathcal{L}_{\cos}$ still grows until 200 epochs. Therefore, the combination of geometric losses in the RGB setting provides our best configuration. In the XYZ setting, however, the combination of geometric loss terms provides the best result, even after the 150 epoch schedule.



\begin{table}[h]
\begin{center}
\begin{tabular}{l|c|c|c|c}
\hline
    \begin{tabular}[c]{@{}c@{}}Models (XYZ)\end{tabular} & \begin{tabular}[c]{@{}c@{}}ACC\\ (0.05)\end{tabular} & \begin{tabular}[c]{@{}c@{}}ACC\\ (0.10)\end{tabular} & EPE & \begin{tabular}[c]{@{}c@{}}ADE\end{tabular} \\ \hline
    F3D & 23.71\% & 56.05\% & 0.1705 & 22.83 \\ 
    F3D + $\mathcal{L}_{pp}$ & 27.79\% & 60.06\% & 0.1567 & 21.96 \\ 
    F3D + $\mathcal{L}_{\cos}$ & 25.30\% & 58.15\% & 0.1615 & 21.17 \\ 
    F3D + $\mathcal{L}_{pp}$ + $\mathcal{L}_{\cos}$ & \textbf{28.22}\% & \textbf{60.11}\% & \textbf{0.1556} & \textbf{20.75} \\ \hline
    F3D + $\mathcal{L}_{pp}$ + $\mathcal{L}_{\cos}$ & 28.50\% & 60.39\% & 0.1553 & 20.77 \\ \hline
    \end{tabular}
\end{center}
\caption{Ablation study for FlowNet3D geometry-only model. The last row is the result trained for 200 epochs. Other rows including the original FlowNet3D are all trained for 150 epochs.}
\label{tab:ablation_xyz}
\end{table}

\begin{table}[h]
\begin{center}
\begin{tabular}{l|c|c|c|c}
\hline
    \begin{tabular}[c]{@{}c@{}}Models (RGB)\end{tabular} & \begin{tabular}[c]{@{}c@{}}ACC\\ (0.05)\end{tabular} & \begin{tabular}[c]{@{}c@{}}ACC\\ (0.10)\end{tabular} & EPE & \begin{tabular}[c]{@{}c@{}}ADE\end{tabular} \\ \hline
    F3D & 25.37\% & 57.85\% & 0.1694 & 22.58 \\
    F3D + $\mathcal{L}_{pp}$ & \textbf{28.52}\% & \textbf{62.75}\% & \textbf{0.1391} & 21.74 \\
    F3D + $\mathcal{L}_{\cos}$ & 26.84\% & 61.57\% & 0.1454 & \textbf{20.96} \\
    F3D + $\mathcal{L}_{pp}$ + $\mathcal{L}_{\cos}$ & 26.05\% & 60.53\% & 0.1492 & 21.27 \\ \hline
    F3D + $\mathcal{L}_{pp}$ + $\mathcal{L}_{\cos}$ & 30.33\% & 63.43\% & 0.1369 & 21.14 \\ \hline
    \end{tabular}
\end{center}
\caption{Ablation study for FlowNet3D colour model. The last row is the result trained for 200 epochs. Other rows including the original FlowNet3D are all trained for 150 epochs.}
\label{tab:ablation_rgb}
\end{table}

\section{Conclusion}\label{sec:conclusion}
In this paper, we introduced FlowNet3D++, which to the best of our knowledge is the state-of-art point cloud-based deep scene flow estimator. We contribute two principled geometric constraints that each improve the accuracy of the state-of-art of point cloud based deep scene flow from \textbf{57.85}\% to \textbf{63.43}\%. We also contribute a novel geometric based scene flow benchmark pipeline in dynamic reconstruction context. Within our deformable scene flow benchmark, our FlowNet3D++ achieves up to 15.0\% less reconstruction error than FlowNet3D, and up to a 35.2\% improvement over KillingFusion alone. 

\section*{Acknowledgements}
We gratefully acknowledge the European Commission Project Multiple-actOrs Virtual Empathic
CARegiver for the Elder (MoveCare) grant for financially supporting the authors of this
work.



{\small
\bibliographystyle{ieee}
\bibliography{922}
}



\section*{Supplementary material}

# FlowNet3D++: Geometric Losses For Deep Scene Flow Estimation
## (Supplementary Material)

Zirui Wang
ryan@robots.ox.ac.uk

Shuda Li
shuda@robots.ox.ac.uk

Henry Howard-Jenkins
henryhj@robots.ox.ac.uk

Victor Adrian Prisacariu
victor@robots.ox.ac.uk

Min Chen
min.chen@oerc.ox.ac.uk

## 1. Further FlowNet3D++ Results With Our Dynamic Reconstruction Benchmark

We present more qualitative results to demonstrate that integrating FlowNet3D++ is useful for high-level dynamic reconstruction applications. We utilise our dynamic reconstruction benchmark pipeline to make qualitative comparisons between dynamic reconstruction results with FlowNet3D++ integrated and without FlowNet3D++. For dynamic reconstruction baseline, we implement SobolevFusion[4], which is an improved version of KillingFusion[3].

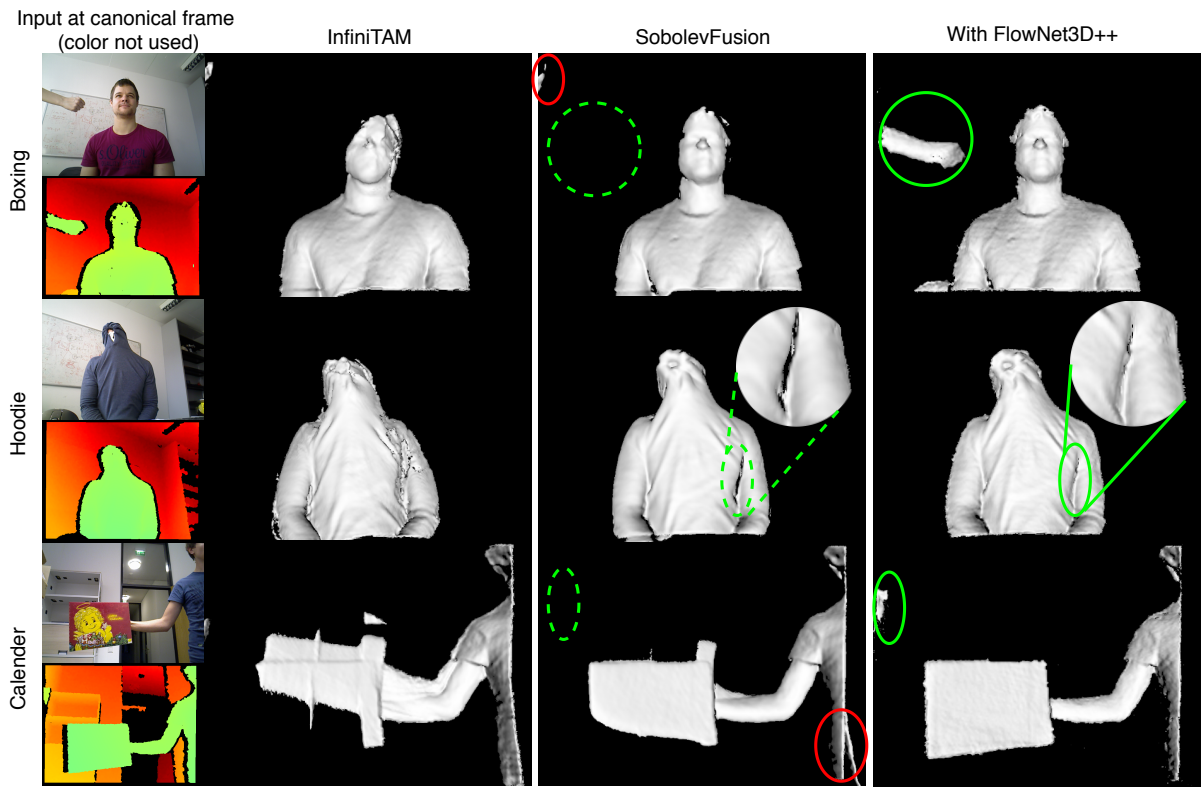

Figure 1. Results of running our benchmark on VolumeDeform [1] Dataset. Aside from obvious differences, we use red circles to highlight places where integrating FlowNet3D++ produces less distortion and green circles to highlight places integrating FlowNet3D++ preserves more detail. Note there is no foreground mask, the background is missing because of volume size. For scenes with more rigid components (the hoodie sequence), we have similar results with SobolevFusion because the rigid tracker lifts the most weight.

1

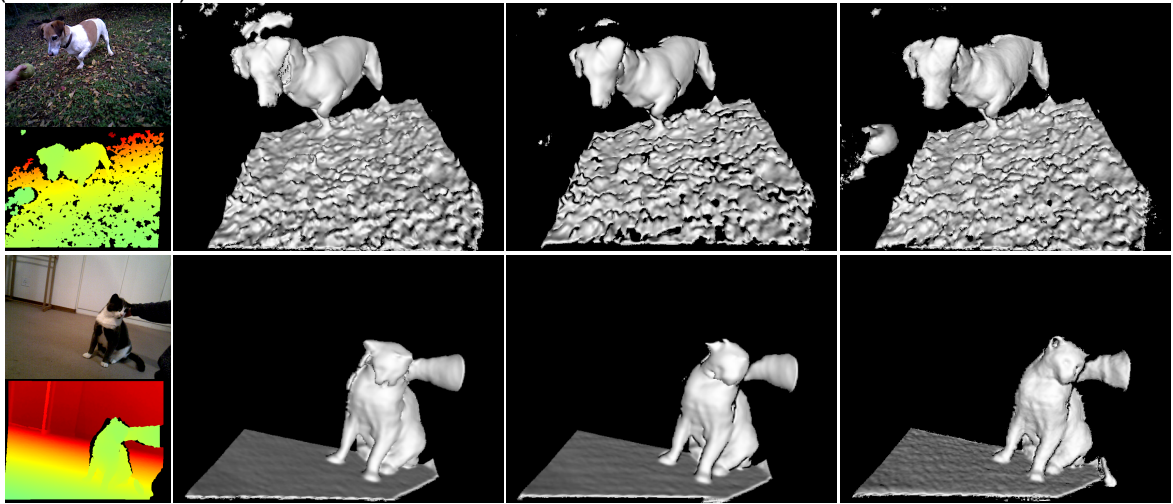

Figure 2. Results of running our benchmark on a dataset used in [2]. Our results have fewer artefacts and more details than SobolevFusion after few hundreds of frames.

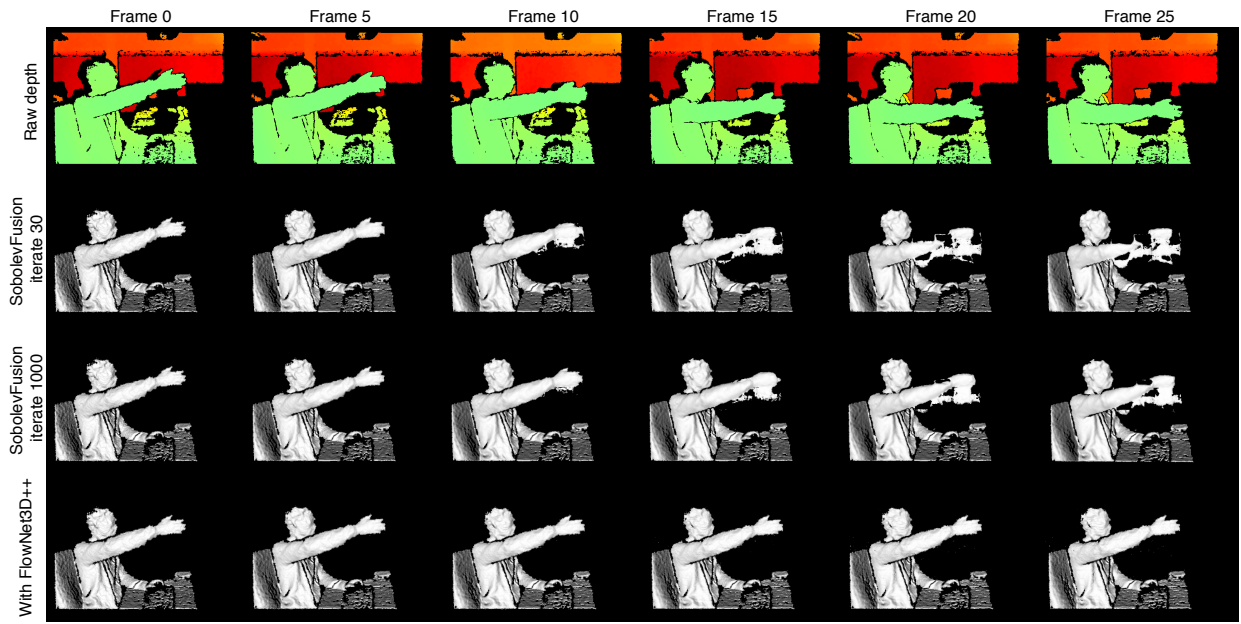

Figure 3. A comparison of the different number of iterations on a video we recorded. We show the canonical model at different time stamps. From the figure we can see that when the iterations increase, the SobolevFusion results improve. However, it is still difficult to register the live volume to the canonical model. In contrast, integrating FlowNet3D++ enables the handling of large motion.

2

## 2. FlowNet3D++ Non-Rigid Registration

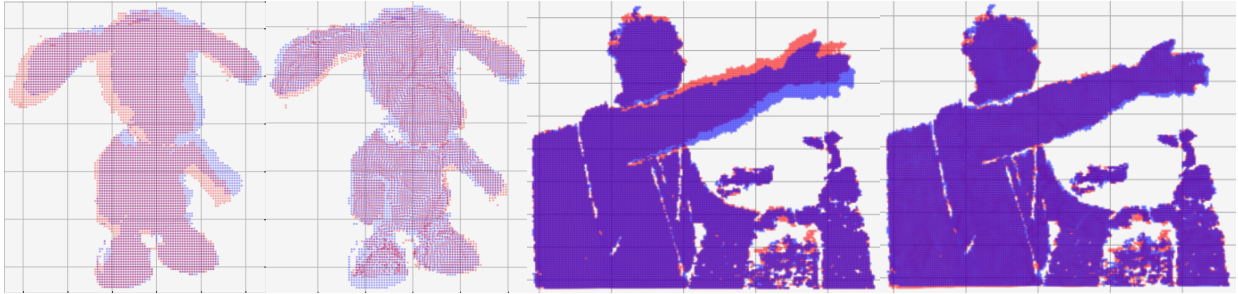

Figure 4. Point clouds warped using the deep scene flow. The left and right pairs are the results using the *Snoopy* sequence and *Waving* sequence, respectively. It can be seen that FlowNet3D++ has aligned the source point cloud (blue) with the target point cloud (red) effectively. Note that although the colour scheme for each of the pairs appears different, this is simply caused by the differing densities of their respective point clouds.



\end{document}